\documentclass[11pt,a4paper]{article}
\usepackage[hyperref]{eacl2021}
\usepackage{times}
\usepackage{latexsym}

\usepackage{microtype}
\usepackage{graphicx}
\usepackage{bbm}
\usepackage{multirow}
\usepackage{amsmath}
\usepackage{subfig}
\usepackage{arydshln}
\usepackage{tipa}
\aclfinalcopy 

\newcommand{\zd}[1]{\textcolor{orange}{\bf\small [#1 --ZD]}}
\newcommand{\gn}[1]{\textcolor{magenta}{\bf\small [#1 --GN]}}

\title{Word Alignment by Fine-tuning Embeddings on Parallel Corpora} 

\iffalse
\author{Zi-Yi Dou\textsuperscript{$\diamondsuit \heartsuit$}  \ \  Graham Neubig\textsuperscript{$\clubsuit$} \\
\textsuperscript{{$\diamondsuit$}}Information Sciences Institute, University of Southern California \\ \textsuperscript{{$\heartsuit$}}University of California, Los Angeles  \\ \textsuperscript{{$\clubsuit$}}Language Technologies Institute, Carnegie Mellon University \\
  {\tt ziyidou@isi.edu \ \ gneubig@cs.cmu.edu}}
\else
\author{Zi-Yi Dou, Graham Neubig\\
  Language Technologies Institute, Carnegie Mellon University \\
  {\tt \{zdou,gneubig\}@cs.cmu.edu }}
\fi

\date{}

\begin{document}
\maketitle
\begin{abstract}
Word alignment over parallel corpora has a wide variety of applications, including learning translation lexicons, cross-lingual transfer of language processing tools, and automatic evaluation or analysis of translation outputs. The great majority of past work on word alignment has worked by performing unsupervised learning on parallel text. Recently, however, other work has demonstrated that pre-trained contextualized word embeddings derived from multilingually trained language models (LMs) prove an attractive alternative, achieving competitive results on the word alignment task even in the absence of explicit training on parallel data. In this paper, we examine methods to marry the two approaches: leveraging pre-trained LMs but fine-tuning them on parallel text with objectives designed to improve alignment quality, and proposing methods to effectively extract alignments from these fine-tuned models. We perform experiments on five language pairs and demonstrate that our model can consistently outperform previous state-of-the-art models of all varieties. In addition, we demonstrate that we are able to train multilingual word aligners that can obtain robust performance on different language pairs.
Our aligner, \textbf{\textsc{AWESoME}} (\textbf{\textsc{A}}ligning \textbf{\textsc{W}}ord \textbf{\textsc{E}}mbedding \textbf{\textsc{S}}paces \textbf{\textsc{o}}f \textbf{\textsc{M}}ultilingual \textbf{\textsc{E}}ncoders), with pre-trained models is available at \url{https://github.com/neulab/awesome-align}. 
\end{abstract}

\section{Introduction}

Word alignment is a useful tool to tackle a variety of natural language processing (NLP) tasks, including learning translation lexicons~\citep{ammar2016massively,cao2019multilingual}, cross-lingual transfer of language processing tools~\citep{yarowsky2001inducing,pado2009cross,tiedemann-2014-rediscovering,agic2016multilingual,mayhew2017cheap,nicolai-yarowsky-2019-learning}, semantic parsing~\citep{herzig2018decoupling} and speech recognition~\citep{xu2019improving}.
In particular, word alignment plays a crucial role in many machine translation (MT) related methods, including guiding learned attention~\citep{liu2016neural}, incorporating lexicons during decoding~\citep{arthur2016incorporating}, domain adaptation~\citep{hu2019domain}, unsupervised MT~\citep{ren2020retrieve} and automatic evaluation or analysis of translation models~\citep{bau2018identifying,stanovsky2019evaluating,neubig2019compare,wang2020inference}.
However, with neural networks advancing the state of the arts in almost every field of NLP, tools developed based on the 30-year-old IBM word-based translation models~\citep{brown1993mathematics}, such as GIZA++~\citep{och2003systematic} or fast-align~\citep{dyer2013simple},  remain popular choices for word alignment tasks.

\begin{figure}[t]
\centering
\includegraphics[width=0.47\textwidth]{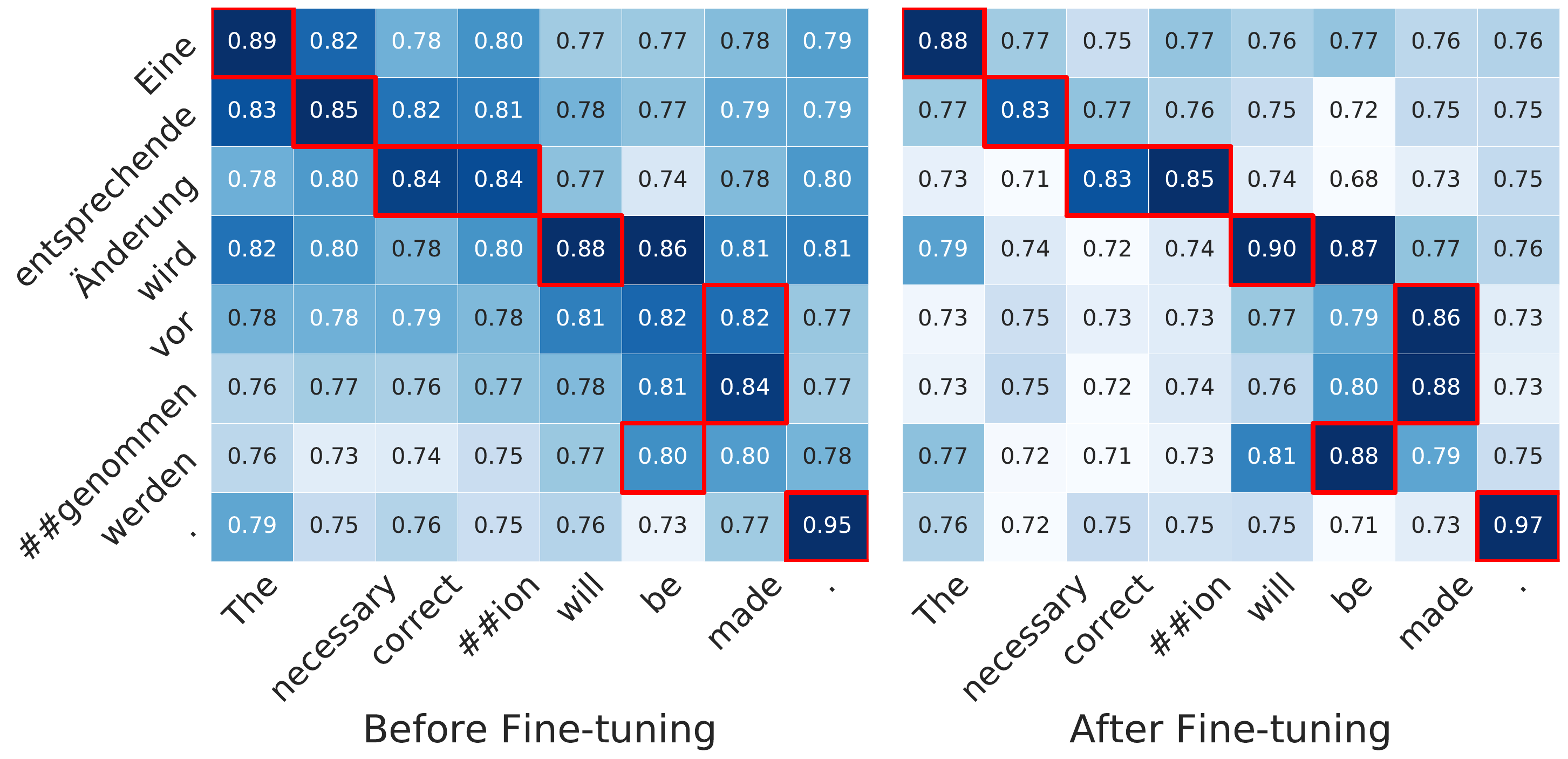}
\caption{Cosine similarities between subword representations in a parallel sentence pair before and after fine-tuning. Red boxes indicate the gold alignments.} 
\label{fig:1}
\end{figure}

One alternative to using statistical word-based translation models to learn alignments would be to instead train state-of-the-art neural machine translation (NMT) models on parallel corpora, and extract alignments therefrom, as examined by \citet{luong2015effective,garg2019jointly,zenkel2020end}.
However, these methods have two disadvantages (also shared with more traditional alignment methods): (1) they are directional and the source and target side are treated differently and (2) they cannot easily take advantage of large-scale contextualized word embeddings derived from language models (LMs) multilingually trained on monolingual corpora~\citep{Devlin2019BERTPO,lample2019cross,conneau2019unsupervised}, which have proven useful in other cross-lingual transfer settings~\citep{libovicky2019language,hu20icml}.
In the field of word alignment,~\citet{sabet2020simalign} have recently proposed methods to align words using multilingual contextualized embeddings and achieve good performance even in the absence of explicit training on parallel data, suggesting that these are an attractive alternative for neural word alignment. 

In this paper, we investigate if we can combine the best of the two lines of approaches.
Concretely, we leverage pre-trained LMs and fine-tune them on parallel text with not only LM-based objectives, but also unsupervised objectives over the parallel corpus designed to improve alignment quality.
Specifically, we propose a self-training objective, which encourages aligned words to have further closer contextualized representations, and a parallel sentence identification objective, which enables the model to bring parallel sentences' representations closer to each other.
In addition, we propose to effectively extract alignments from these fine-tuned models using probability thresholding or optimal transport.


We perform experiments on five different language pairs and demonstrate that our model can achieve state-of-the-art performance on all of them.
In analysis, we find that these approaches also generate more aligned contextualized representations after fine-tuning (see Figure~\ref{fig:1} as an example) and we can incorporate supervised signals within our paradigm.
Importantly, we show that it is possible to train multilingual word aligners that can obtain robust performance even in zero-shot settings, making them a valuable tool that can be used out-of-the-box with good performance over a wide variety of language pairs.

\section{Methods}
Formally, the task of word alignment can be defined as: given a sentence $\mathbf{x}= \langle x_1, \cdots, x_n \rangle $ in the source language and its corresponding parallel sentence $\mathbf{y}= \langle y_1, \cdots, y_m \rangle $ in the target language, a word aligner needs to find a set of pairs of source and target words:
$$
A=\{ \langle x_i, y_j\rangle: x_i \in \mathbf{x}, y_j \in \mathbf{y} \},
$$
where for each word pair $\langle x_i, y_j \rangle$, $x_i$ and $y_j$ are semantically similar to each other within the context of the sentence. 

In the following paragraphs, we will first illustrate how we extract alignments from contextualized word embeddings, then describe our objectives designed to improve alignment quality.

\subsection{Extracting Alignments from Embeddings}
\label{sec:extract}


Contextualized word embedding models such as BERT \citep{Devlin2019BERTPO} and RoBERTa \citep{Liu2019RoBERTaAR} represent words using continuous vectors calculated in context, and have achieved impressive performance on a diverse array of NLP tasks. {Multilingually trained word embedding models such as multilingual BERT can generate contextualized embeddings across different languages.}
These models can be used to extract contextualized word embeddings  $h_{\mathbf{x}} = \langle h_{x_1}, \cdots, h_{x_n} \rangle $ and $h_{\mathbf{y}}=\langle h_{y_1}, \cdots, h_{y_m} \rangle $ for each pair of parallel sentences $\mathbf{x}$ and $\mathbf{y}$.
Specifically, this is done by extracting the hidden states of the $i$-th layer of the model, where $i$ is an empirically-chosen hyper-parameter.
{Given these contextualized word embeddings, we propose two methods to calculate unidirectional alignment scores based on probability simplexes and optimal transport. We then turn these alignment scores into alignment matrices and reconcile alignments in the forward and backward directions.} 

\begin{figure*}[t]
\centering
\includegraphics[width=1.0\textwidth]{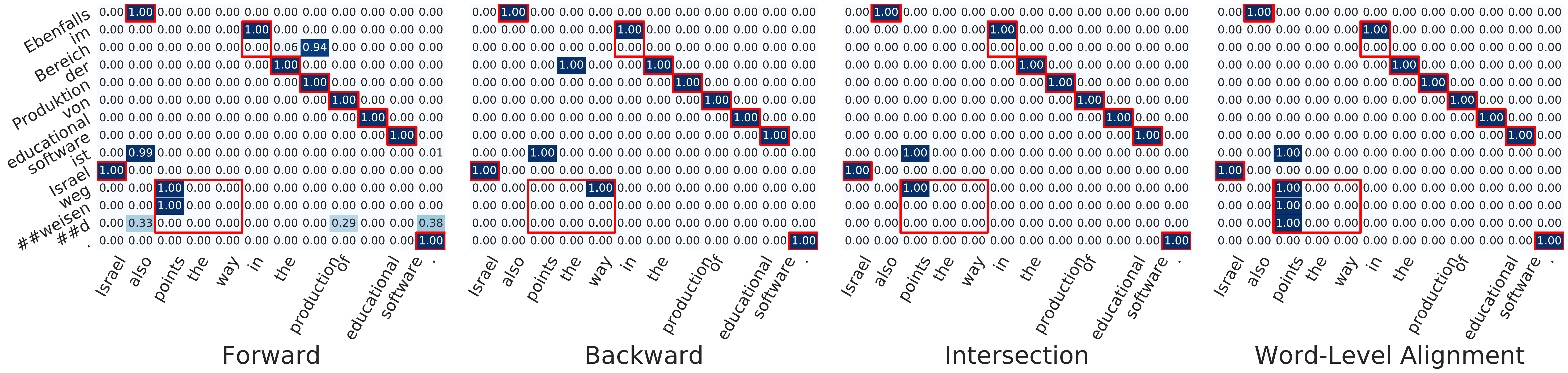}
\caption{Extracting word alignments from multilingual BERT using probability thresholding (\textit{softmax}). Red boxes denote the gold alignments.} 
\label{fig:example1}
\end{figure*}

\paragraph{Probability Thresholding. }
In this method, for each word in the source/target sentence, we calculate a value on the probability simplex for each word in the aligned target/source sentence, and then select all values that exceed a particular threshold as ``aligned'' words.
Concretely, taking inspiration from attention mechanisms~\citep{Bahdanau2015NeuralMT,vaswani2017attention}, we take the contextualized embeddings $h_{\mathbf{x}}$ and $h_{\mathbf{y}}$ and compute the dot products between them and get the similarity matrix:
$$
S = h_{\mathbf{x}} h_{\mathbf{y}}^T.
$$

Then, we apply a normalization function $\mathcal{N}$ to convert the similarity matrix into values on the probability simplex {$S_{\mathbf{xy}} = \mathcal{N}(S)$, and treat $S_{\mathbf{xy}}$ as the source-to-target alignment matrix.} In this paper, we propose to use {\textit{softmax}} and a sparse variant $\alpha$-entmax 
~\citep{peters2019sparse} to do the normalization. Compared with the \textit{softmax} function, $\alpha$-entmax can produce sparse alignments for any $\alpha > 1$ and assign non-zero probability
to a short list of plausible word pairs, where a higher $\alpha$ will lead to a more sparse alignment. 

\paragraph{Optimal Transport.}
The goal of optimal transport~\citep{monge1781memoire,cuturi2013sinkhorn} is to find a mapping that moves probability from one distribution to another, which can be used to find an optimal matching of similar words between two sequences~\citep{kusner2015word}. 
Formally, in a discrete optimal transport problem, we are given two point sets $\{ {x_i} \}_{i=1}^n$ and $\{ {y_j} \}_{j=1}^m$ associated with their probability distributions $p_{\mathbf{x}}$ and $p_{\mathbf{y}}$ where $\sum_i p_{x_i} =1$ and $\sum_j p_{y_j} = 1$. Also, a function $C({x_i}, {y_j})$ defines the cost of moving point ${x_i}$ to ${y_j}$. The goal of optimal transport is to find a
mapping that moves probability mass from $\{ {x_i} \}_{i=1}^n$ to $\{ {y_j} \}_{j=1}^m$ and
the total cost of moving the mass between points is
minimized. In other words, it finds the transition matrix $S_{\mathbf{xy}}$ that minimizes:
\begin{equation}
\label{eqn:ot}
    \sum_{i, j} C({x_i}, {y_j}) {S_{\mathbf{xy}}}_{ij},
\end{equation}
where $S_{\mathbf{xy}}\mathbf{1}_m = p_{\mathbf{x}}$ and $S_{\mathbf{xy}}^T\mathbf{1}_n = p_{\mathbf{y}}$.
The resulting transition matrix is self-normalized and sparse~\citep{swanson2020rationalizing}, making it appealing alternative towards extracting alignments from word embeddings.

In this paper, we propose to adapt optimal transport techniques to the task of word alignment. Concretely, we treat the parallel sentences $\mathbf{x}$ and $\mathbf{y}$ as two point sets and assume each word is uniformly distributed. The cost function is obtained by computing the pairwise distance (e.g. cosine distance) between $h_{\mathbf{x}}$ and $h_{\mathbf{y}}$, and all the distance values are scaled to [0, 1] with min-max normalization. The optimal transition matrix ${S_{\mathbf{xy}}}$ to Equation \ref{eqn:ot} can be calculated using the Sinkhorn-Knopp matrix scaling algorithm~\citep{sinkhorn1967concerning}. If the value of ${S_{\mathbf{xy}}}_{ij}$ is high, ${x_i}$ and ${y_j}$ are likely to have similar semantics and values that exceed a particular threshold will be considered as ``aligned''.




\paragraph{Extracting Bidirectional Alignments.} After we obtain both the source-to-target and target-to-source alignment probability matrices $S_{\mathbf{xy}}$ and $S_{\mathbf{yx}}$ using the previous methods, we can deduce the final alignment matrix by taking the intersection of the two matrices:
$$
A = (S_{\mathbf{xy}} > c) * (S_{\mathbf{yx}}^T> c), 
$$
where $c$ is a threshold and $A_{ij} = 1$ means $x_i$ and $y_j$ are aligned.

Note that growing heuristics such as {\it grow-diag-final}~\citep{och2000improved,koehn2005edinburgh} that are popular in statistical word aligners can also be applied in our alignment extraction algorithms, and we will demonstrate the effect of these heuristics in the experiment section.

\paragraph{Handling Subwords.}
Subword segmentation techniques~\citep{sennrich2016neural,kudo2018sentencepiece} are widely used in training LMs, thus the above alignment extraction methods can only produce alignments on the subword level. To convert them to word alignments, we follow previous work~\citep{sabet2020simalign,zenkel2020end} and consider two words to be aligned if any of their subwords are aligned. {Figure~\ref{fig:example1} shows a concrete example of how we extract word-level alignments from a pre-trained embedding model. }

\subsection{Fine-tuning Contextualized Embeddings for Word Alignment}
While language models can be used to produce reasonable word alignments even without any fine-tuning~\citep{sabet2020simalign}, we propose objectives that further improve their alignment ability if we have access to parallel data.

\paragraph{Masked Language Modeling (MLM).} {\citet{gururangan-etal-2020-dont} suggest that we can gain improvements in downstream tasks by further pre-training LMs on the task datasets. Therefore, we propose to fine-tune the LMs with a masked language modeling objective on both the source and target side of parallel corpora.} 
Specifically, given a pair of parallel sentences $\mathbf{x}$ and $\mathbf{y}$, we choose 15\% of the token positions randomly for both $\mathbf{x}$ and $\mathbf{y}$, and for each chosen token, we replace it with (1) the {\tt [MASK]} token 80\% of the time (2) a random token 10\% of the time and (3) unchanged 10\% of the time. The model is trained to reconstruct the original tokens given the masked sentences $\mathbf{x}^{mask}$ and $\mathbf{y}^{mask}$:
\begin{equation}
\label{eqn:mlm}
L_{MLM}= \log p(\mathbf{x}|\mathbf{x}^{mask}) + \log p(\mathbf{y}|\mathbf{y}^{mask}).
\end{equation}

\paragraph{Translation Language Modeling (TLM).}
{The MLM objective only requires monolingual data and the model cannot make direct connections between parallel sentences. To solve the issue, }similarly to~\citet{lample2019cross}, we concatenate parallel sentences $\mathbf{x}$ and $\mathbf{y}$ and perform MLM on the concatenated data. Compared with MLM, the translation language modeling (TLM) objective enable the model to align the source and target representations. Different from~\citet{lample2019cross}, we feed source and target sentences twice in different orders instead of resetting the positions of target sentences:
\begin{equation}
\label{eqn:tlm}
\begin{aligned}
L_{TLM} & = \log p([\mathbf{x}; \mathbf{y}] |[\mathbf{x}^{mask}; \mathbf{y}^{mask}]) \\
& + \log p([\mathbf{y}; \mathbf{x}] |[\mathbf{y}^{mask}; \mathbf{x}^{mask}]) .
\end{aligned}
\end{equation}

\paragraph{Self-training Objective (SO).}
We also propose a self-training objective for fine-tuning LMs which is similar to the EM algorithm used in the IBM models and the agreement constraints in~\citet{tamura2014recurrent}. Specifically, at each training step, we first use our alignment extraction methods (described in Section~\ref{sec:extract}) to extract the alignment $A$ for $\mathbf{x}$ and $\mathbf{y}$, then maximize the following objective:
\begin{equation}
\label{eqn:so}
    L_{SO}=\sum_{i,j} A_{ij} \frac{1}{2} ( \frac{S_{{\mathbf{xy}}_{ij}}}{n} + \frac{S^{\mathrm{T}}_{{\mathbf{yx}}_{ij}}}{m} ).
\end{equation}

Intuitively, this objective encourages words aligned in the first pass of alignment to have further closer contextualized representations. In addition, because of the intersection operation during extraction, the self-training objective can ideally reduce spurious alignments and encourage the source-to-target and target-to-source alignments to be symmetrical to each other by exploiting their agreement~\citep{liang-etal-2006-alignment}.

\paragraph{Parallel Sentence Identification (PSI).}
We also propose a contrastive parallel sentence identification loss that attempts to make parallel sentences more similar than mismatched sentence pairs~\citep{liu2015contrastive,legrand2016neural}.
This encourages the overall alignments of embeddings on both word and sentence level to be closer together. Concretely, we randomly select a pair of parallel or non-parallel sentences $\langle\mathbf{x}', \mathbf{y}'\rangle$ from the training data with equal probability. Then, the model is required to predict whether the two sampled sentences are parallel or not. The representation of the first {\tt [CLS]} token is fed into a multi-layer perceptron to output a prediction score $s(  \mathbf{x}', \mathbf{y}' )$. Denoting the binary label as $l$, the objective function can be written as:
\begin{equation}
\label{eqn:psi}
L_{PSI} = l\log s(\mathbf{x}', \mathbf{y}') + (1-l) \log (1-s(\mathbf{x}', \mathbf{y}')).
\end{equation}

\paragraph{Consistency Optimization (CO).} While the self-training objective can potentially improve the symmetricity between forward and backward alignments, following previous work on machine translation and multilingual representation learning~\citep{cohn2016incorporating,zhang2019regularizing,hu2020explicit}, we use an objective to explicitly encourage the consistency between the two alignment matrices. Specifically, we maximize the trace of $S_{\mathbf{xy}}^\mathrm{T}S_{\mathbf{yx}}$:
\begin{equation}
    L_{CO} = \frac{\text{trace}(S_{\mathbf{xy}}^\mathrm{T}S_{\mathbf{yx}})}{\min(m, n)}.
\end{equation}

\paragraph{Our Final Objective.} In summary, our training objective is a combination of the proposed objectives and we train the model with them jointly at each training step:
$$
L = L_{MLM}+L_{TLM}+L_{SO} +L_{PSI} + \beta L_{CO},
$$
where $\beta$ is set to 0 or 1 in our experiments.

\begin{table}[t]
  \centering
   \resizebox{0.5\textwidth}{!}{%
  \begin{tabular}{llllll}
&  \multicolumn{1}{c}{ \bf De-En} & \bf Fr-En & \bf Ro-En & \bf Ja-En & \bf Zh-En   \\
   \hline
  \hline
  \#Train Sents. & 1.9M & 1.1M & 450K & 444K & 40K \\
  \#Test Sents. & 508 & 447 & 248 & 582 & 450\\ 
    \end{tabular}
    }
    \caption{\label{tab:stats}Statistics of datasets.}
  \end{table}
  
\begin{table*}[ht]
  \centering
   \resizebox{0.8\textwidth}{!}{%
  \begin{tabular}{lllllll}
 \multicolumn{1}{l}{ \bf Model }& \bf Setting &\multicolumn{1}{c}{ \bf De-En} & \bf Fr-En & \bf Ro-En & \bf Ja-En & \bf Zh-En   \\
   \hline
  \hline
   \multicolumn{7}{l}{ \it Baseline} \\
   \hline
     \multicolumn{1}{l}{ SimAlign}& \it w/o fine-tuning & 18.8 & 7.6 & 27.2 & 46.6 & 21.6  \\
   \multicolumn{1}{l}{ fast\_align} & \it bilingual & 27.0 & 10.5 & 32.1 & 51.1 & 38.1\\
   \multicolumn{1}{l}{ eflomal} & \it bilingual & 22.6 & 8.2 & 25.1 & 47.5 & 28.7 \\
  \multicolumn{1}{l}{ GIZA++ }& \it bilingual  & 20.6 & 5.9 & 26.4 & 48.0 & 35.1  \\
  \multicolumn{1}{l}{ \citet{zenkel2020end} }&  \it bilingual &16.0 & 5.0 & 23.4 & - & - \\
  \multicolumn{1}{l}{ \citet{chen2020accurate}}&  \it bilingual &15.4 & 4.7 & 21.2 & - & - \\
  \hline
  \hline
  \multicolumn{7}{l}{ \it Ours} \\
  \hline
  \multirow{4}{*}{$\alpha$-entmax}
  & \it w/o fine-tuning  & 18.1 & 5.6 & 29.0 & 46.3 & 18.4\\
  & \it bilingual & 16.1 &   \textbf{\textit{4.1}} & 23.4 & 38.6 & 15.4\\
  & \it multilingual ($\beta$ = 0) &  15.4 & \textbf{\textit{4.1}} &  22.9 & \textbf{\textit{37.4}} & \bf 13.9 \\
  & \it multilingual ($\beta$ = 1)&  \textbf{\textit{15.0}} & 4.5 & \bf 20.8 & 38.7 & 14.5 \\
  & \it zero-shot & 16.0 & 4.3 & 28.4 & 44.0 & \bf 13.9 \\
  \hline
   \multirow{4}{*}{\it softmax}
 & \it w/o fine-tuning &  17.4 & 5.6 & 27.9 & 45.6 & 18.1 \\
  & \it bilingual & 15.6 & \bf 4.4 &  23.0 &  38.4 &  15.3\\
  & \it multilingual ($\beta$ = 0) &  {{15.3}} & \bf 4.4 & 22.6 & \bf 37.9 & \textbf{\textit{13.6}} \\
  & \it multilingual ($\beta$ = 1) &  \textbf{{15.1}} & 4.5 & \textbf{\textit{20.7}} & 38.4 & 14.5 \\
  & \it zero-shot & 15.7 & 4.6 & 27.2 & 43.7 & 14.0 \\
    \end{tabular}
    }
    \caption{\label{tab:main} Performance (AER) of our models in bilingual, multilingual and zero-shot settings. The best scores for each alignment extraction method are in {\bf bold} and the overall best scores are in \textbf{\textit{italicized bold}}.} 
  \end{table*}

\section{Experiments}
In this section, we first present our main results, then conduct several ablation studies and analyses of our models.

\subsection{Setup}

\paragraph{Datasets.}
We perform experiments on five different language pairs, namely  German-English (De-En), French-English (Fr-En), Romanian-English (Ro-En), Japanese-English (Ja-En) and Chinese-English (Zh-En). 
For the De-En, Fr-En, Ro-En datasets, we follow the experimental setting of previous work ~\citep{zenkel2019adding,garg2019jointly,zenkel2020end}. The training and test data for Ro-En and Fr-En are provided by~\citet{mihalcea2003evaluation}. The Ro-En training data are also augmented by the Europarl v8 corpus~\citep{koehn2005europarl}. For the De-En data, the Europarl v7 corpus is used as training data and the gold alignments are provided by~\citet{vilar2006aer}. The Ja-En dataset is obtained from the Kyoto Free Translation Task (KFTT) word alignment data~\citep{neubig11kftt}, and the Japanese sentences are tokenized with the KyTea tokenizer~\citep{neubig2011pointwise}. The Zh-En dataset is obtained from the TsinghuaAligner website\footnote{\url{http://nlp.csai.tsinghua.edu.cn/~ly/systems/TsinghuaAligner/TsinghuaAligner.html}}. We treat their evaluation set as the training data and use the test set in~\citet{liu2015contrastive} ignoring possible alignments. The De-En, En-Fr datasets contain the distinction between sure and possible alignment links. The statistics of these datasets are shown in Table~\ref{tab:stats}. We use the Ja-En development set to tune the hyper-parameters.

\paragraph{Baselines.}
We compare our models with:
\begin{itemize}
    \item fast\_align~\citep{dyer2013simple}: a popular statistical word aligner which is a simple, fast reparameterization of IBM Model 2. 
    \item eflomal~\citep{ostling2016efficient}: an efficient statistical word aligner using a Bayesian model with Markov Chain Monte Carlo (MCMC) inference.
    \item GIZA++~\citep{och2003systematic,gao2008parallel}: an implementation of IBM models. Following previous work~\citep{zenkel2020end}, we use five iterations each for Model 1, the HMM model, Model 3 and Model 4.
    \item SimAlign~\citep{sabet2020simalign}: a BERT-based word aligner that is not fine-tuned on any parallel data. The authors propose three alignment extraction methods and we implement their IterMax model with default parameters.
    \item \citet{zenkel2020end} and~\citet{chen2020accurate}: two state-of-the-art neural word aligners based on MT models. 
\end{itemize}

\paragraph{Implementation Details.} Our main results are obtained by using the probability thresholding method on the contextualized embeddings in the 8-th layer of multilingual BERT-Base (mBERT; \citet{Devlin2019BERTPO}) and we will discuss this choice in our ablation studies. We use the AdamW optimizer~\citep{Loshchilov2019DecoupledWD} with a learning rate of 2e-5 and the batch size is set to 8. Following~\citet{peters2019sparse}, we set $\alpha$ to 1.5 for $\alpha$-entmax. The threshold $c$ is set to 0 for $\alpha$-entmax and 0.001 for \textit{softmax} and optimal transport. Unless otherwise stated, $\beta$ is set to 0. We mainly evaluate the model performance using Alignment Error Rate (AER).

\subsection{Main Results}
We first train our model on each individual language pair, then investigate if it is possible to train multilingual word aligners.

\paragraph{Bilingual Model Performance.} From Table~\ref{tab:main}, we can see that our \textit{softmax} model can achieve consistent improvements over the baseline models, demonstrating the effectiveness of our proposed method. Surprisingly, directly extracting alignments from mBERT (the \textit{w/o fine-tuning} setting) can already achieve better performance than the popular statistical word aligner GIZA++ on 4 out of 5 settings, especially in the Zh-En setting where the size of parallel data is small. 

\paragraph{Multilingual Model Performance.} We also randomly sample 200k parallel sentence pairs from each language pair (except for Zh-En where we take all of its 40k parallel sentences) and concatenate them together to train multilingual word aligners. As shown in Table~\ref{tab:main}, the multilingually trained word aligners can achieve further improvements and they consistently outperform our bilingual word aligners and all the baselines even though the size of training data for each individual language pair is smaller. The results demonstrate that we can indeed obtain a neural word aligner that has state-of-the-art and robust performance across different language pairs. We also test the performance of our consistency optimization objective in this setting. We can see that incorporating this objective ($\beta$=1) can significantly improve the model performance on Ro-En, while it also deteriorates the Ja-En and Zh-En performance by a non-negligible margin. We find that this is because the CO objective can significantly improve the alignment recall while sacrificing the precisions, and our Ro-En dataset tends to favor models with high recall and the Ja-En and Zh-En datasets have an opposite tendency.

\paragraph{Zero-Shot Performance.} In this paragraph, we want to find out how our models perform on language pairs that it has never seen during training. To this end, for each language pair, we train our model with data of all the other language pairs and test its performance on the target language pair. Results in Table~\ref{tab:main} demonstrate that training our models with parallel data on \emph{other} language pairs can still improve the model performance on the target language pair.
This is a very important result, as it indicates that our model can be used as a off-the-shelf tool for multilingual word alignment for any language supported by the underlying embeddings, \emph{regardless of whether parallel data has been used for training or not}.

\subsection{Ablation Studies}
 \begin{table}[t]
  \centering
   \resizebox{0.5\textwidth}{!}{%
  \begin{tabular}{llllllll}
  & \bf Component& \multicolumn{1}{c}{ \bf De-En} & \bf Fr-En & \bf Ro-En & \bf Ja-En & \bf Zh-En  & \bf Speed  \\
   \hline
  \hline
 \multirow{2}{*}{Prob.} & \textit{softmax} & \bf 17.4 & \bf 5.6 & \bf 27.9 & \bf 45.6 & \bf 18.1  & \bf 33.22 \\
 & $\alpha$-entmax & 18.1 &\bf   5.6 & 29.0 & 46.3 & 18.4 & 32.36\\
 \hline
 \multirow{3}{*}{OT} &  Cosine & 24.4 & 15.7 & 33.7 & 54.0 & 31.1 & 3.36 \\
 & Dot Product & 25.4 & 17.1 & 34.1 & 54.2 & 30.9 & 3.82\\
 & Euclidean  & 20.7 & 15.1 & 33.3 & 53.2 & 29.8 & 3.05\\
    \end{tabular}
    }
    \caption{\label{tab:align}Comparisons of probability thresholding (Prob.) and optimal transport (OT) for alignment extraction. We try both \textit{softmax} and $\alpha$-entmax for probability thresholding and different cost functions for optimal transport. We measure both the extraction speed (\#sentences/seconds) and the alignment quality (AER) on five language pairs, namely German-English (De-En), French-English (Fr-En), Romanian-English (Ro-En), Japanese-English (Ja-En), and Chinese-English (Zh-En). The best scores are in {\bf bold}. }
  \end{table}

In this part, we  compare the performance of different alignment extraction methods, pre-trained embedding models and training objectives. 

\paragraph{Alignment Extraction Methods.} We first compare the performance of our two proposed alignment extraction methods, namely the probability thresholding and optimal transport techniques. We use the representations of the 8-th layer of mBERT following~\citet{sabet2020simalign}. 

As shown in Table~\ref{tab:align}, probability thresholding methods can consistently outperform optimal transport by a large margin on the five language pairs. In addition, probability thresholding methods are much faster than optimal transport. \textit{softmax} is marginally better than $\alpha$-entmax, yet one advantage of $\alpha$-entmax is that we do not need to manually set the threshold. Therefore, we use both \textit{softmax} and $\alpha$-entmax to obtain the main results.

\paragraph{Pre-trained Embedding Models.} In this paragraph, we investigate the performance of three different types of pre-trained embedding models, including mBERT, XLM~\citep{lample2019cross} and XLM-R~\citep{conneau2019unsupervised}. For XLM, we have tried its three released models: 1) XLM-15 (MLM) pre-trained with MLM and supports 15 languages; 2) XLM-15 (MLM+TLM) pre-trained with both the MLM and TLM objectives and supports 15 languages; 3) XLM-100 (MLM) pre-trained with MLM and supports 100 languages. We use \textit{softmax} to extract the alignments.

Because XLM-15 does not support Japanese or Romanian, we only report the performance on the three other language pairs in Table~\ref{tab:lm}. We take representations from different layers and report the performance of the best three layers. We can see that while XLM-15 (MLM+TLM) can achieve the best performance on De-En and Fr-En, the best layer is not consistent across language pairs. On the other hand, the optimal configurations for mBERT are consistent across language pairs. In addition, considering mBERT supports many more languages than XLM-15 (MLM+TLM), we will use mBERT in the following sections.

\begin{table}[t]
  \centering
   \resizebox{0.5\textwidth}{!}{%
  \begin{tabular}{llllll}
  \bf Model & \bf Layer &\multicolumn{1}{c}{ \bf De-En} & \bf Fr-En &  \bf Zh-En   \\
   \hline
  \hline
  \multirow{3}{*}{mBERT}
  & 7 & 18.7 & 6.1 & 19.1\\
  & 8 & \bf 17.4 & \bf 5.6 & \textbf{\textit{18.1}}\\
  & 9 & 18.8 & 6.1 & 20.1\\
  \hline
  \multirow{3}{*}{XLM-15 (MLM)} 
  & 4 & 21.1 & 6.8 & \bf 25.3\\
  & 5 & \bf 20.4 & \bf 6.1 & 26.1\\
  & 6 & 23.2 & 7.7 & 33.3\\
  \hline
  \multirow{3}{*}{XLM-15 (MLM+TLM)} 
  & 4 & 16.4 & 4.9 & \bf 18.6\\
  & 5 & \textbf{\textit{ 16.2}} & \bf \textbf{\textit{4.7 }}& 23.7\\
  & 6 & 18.8 & 5.7 & 26.2\\
  \hline
   \multirow{3}{*}{XLM-100 (MLM)} 
   & 7 & 20.5 & 8.5 & 30.8\\
   & 8 & \textbf{{19.8}} &\textbf{{8.2}} & \bf 28.6\\
   & 9 & 19.9 & 8.8 & 29.3\\
   \hline
  \multirow{3}{*}{XLM-R} 
  & 7 & 24.4 & 10.3 & 33.2\\
  & 8 & \bf 23.1 & \bf 9.2  & 30.7 \\
  & 9 & 24.7 & 11.5 & \bf 28.1\\
    \end{tabular}
    }
    \caption{\label{tab:lm} Comparisons of different LMs in terms of AER. We extract alignments using \textit{softmax} and take representations from different layers of LMs. The best scores for each individual model are in {\bf bold} and the overall best scores are in \textbf{\textit{italicized bold}}.  }
  \end{table}
  
\begin{table*}[ht]
  \centering
   \resizebox{0.7\textwidth}{!}{%
  \begin{tabular}{lllllll}
 \multicolumn{1}{l}{ \bf Model }& \bf Objective &\multicolumn{1}{c}{ \bf De-En} & \bf Fr-En & \bf Ro-En & \bf Ja-En & \bf Zh-En   \\
   \hline
  \hline
  \multirow{5}{*}{\it softmax} 
  & \it All &   15.3 & 4.4 &  22.6 & 37.9 &  13.6 \\
  \cdashline{2-7}
    &  \it All w/o MLM & 15.3 & 4.4 & 22.8 & 38.6 & 13.7\\
   &  \it All w/o TLM & 15.5 & 4.7 & 22.9 & 39.7 & 14.0 \\
   & \it All w/o SO & 16.9 & 4.8 & 23.0 & 39.1 & 15.4 \\
  & \it  All w/o PSI &  15.4 & 4.4 & 22.7 & 37.9 & 13.8 \\
  \hline
    \end{tabular}
    }
    \caption{\label{tab:ablation} Ablation studies on our training objectives in multilingual settings.}
  \end{table*}

\paragraph{Training Objectives.} We also conduct ablation studies on each of our training objectives. We can see from Table~\ref{tab:ablation} that the self-training objective can best improve the model performance. Also, the translation language modeling and parallel sentence identification objectives can marginally benefit the model. The masked language modeling objective, on the other hand, cannot always improve the model and can sometimes even deteriorate the model performance, possibly because the TLM objective already provides the model with sufficient supervision signals.

\subsection{Analysis}
We conduct several analyses to better understand our models. Unless otherwise stated, we perform experiments on the \textit{softmax} model using mBERT.

\paragraph{Incorporating Supervised Signals.} We investigate if our models can benefit from supervised signals. If we have access to word-level gold labels for word alignment, we can simply utilize them in our self-training objectives. Specifically, we can set $A_{ij}$ in Equation~\ref{eqn:so} to 1 if and only if they are aligned. In our experimental settings, we have gold labels for all the Zh-En sentences and 653 sentences from the Ja-En development set. Table~\ref{tab:gold} demonstrates that training our models with as few as 653 labeled sentences can dramatically improve the alignment quality, and combining labeled and unlabeled parallel data can further improve the model performance. This analysis demonstrate the generality of our models as they can also be applied in semi-supervised settings.

\begin{table}[t]
  \centering
   \resizebox{0.4\textwidth}{!}{%
  \begin{tabular}{llll}
  \bf Lang. &  \bf  Unsup. & \bf Sup. & \bf Semi-Sup. \\
  \hline
  \hline
   Zh-En & 15.3 & 12.5 & - \\
  Ja-En & 38.4 & 31.6 & 30.0 \\
   \end{tabular}
    }
    \caption{\label{tab:gold} Incorporating supervised word alignment signals into our model can further improve the model performance in terms of AER. }
  \end{table}

\paragraph{Growing Heuristics.} As stated in Section~\ref{sec:extract}, because our alignment extraction methods essentially take the intersection of forward and backward alignments, growing heuristics can also be applied in our settings. The main motivation of growing heuristics is to improve the recall of the resulting alignments. While effective in statistical word aligners, as shown in Table~\ref{tab:grow-diag}, the growing heuristics only improve our alignment extraction method on the vanilla mBERT model in the Ro-En setting while degrading the model performance on all the other language pairs. After fine-tuning, the growing heuristics can only hurt the model performance, possibly because the self-training objective encourages the forward and backward alignments to be symmetrical. Based on these results, we do not adopt the growing heuristics in our models.

\begin{table}[t]
  \centering
   \resizebox{0.5\textwidth}{!}{%
  \begin{tabular}{llllllll}
  \bf Model & \bf Ext. &\multicolumn{1}{c}{ \bf De-En} & \bf Fr-En & \bf Ro-En & \bf Ja-En & \bf Zh-En  \\
   \hline
  \hline
  \multirow{5}{*}{mBERT} 
  & X-En &24.7 & 14.4 & 31.9 & 54.7 & 27.4\\
  & En-X & 22.6 & 12.2 &32.0 & 52.7 & 29.9  \\
  & \textit{softmax} & 17.4 & 5.6 & 27.9 & 45.6 & 18.1 \\
  &  gd & 18.7 & 9.2 & 27.0 & 48.5 & 23.4 \\
  & gd-final & 18.6 & 9.3 & 26.9 & 48.7 & 23.2 \\
  \hline
  \hline
  \multirow{5}{*}{Ours-Multi.} 
  & X-En & 20.2 & 12.9 & 25.4 & 42.1 & 19.3 \\
  & En-X & 18.1 & 9.3 & 25.9 & 41.7 & 23.5 \\
  & \textit{softmax}& 15.3 & 4.4 & 22.6 & 37.9 & 13.6\\
  & gd & 16.3 & 8.1 & 23.1 & 38.2 & 18.3 \\
  & gd-final & 16.5 & 8.3& 23.2 & 38.7 & 18.5 \\
    \end{tabular}
    }
    \caption{\label{tab:grow-diag} The \textit{grow-diag-final} heuristic can only improve our alignment extraction method in the Romanian-English setting without fine-tuning. ``gd'' refers to grow-diag.}
  \end{table}

\begin{table}[t]
  \centering
   \resizebox{0.45\textwidth}{!}{%
  \begin{tabular}{llll}
  \bf Model & \bf Prec. \% &  \bf Rec. \% & \bf F$_1$ \%   \\
   \hline
  \hline
  \multirow{1}{*}{BERT-En (zero-shot)} & 53.1 & 54.3 & 52.7  \\
  \multirow{1}{*}{fast\_align} & 51.5 & 59.8 & 55.2  \\
  \multirow{1}{*}{GIZA++} & 56.5 & 64.1 & 60.0  \\
  \multirow{1}{*}{SimAlign} & 59.9 & 67.6 & 63.5  \\
  \multirow{1}{*}{Ours} & \bf 60.6 & \bf 68.5 & \bf 64.3  \\
    \end{tabular}
    }
    \caption{\label{tab:ner} Our model is also effective in an annotation projection setting where we train a BERT-based NER model on English data and test it on Spanish data. The best scores are in \textbf{bold}.}
  \end{table}

\begin{table*}[ht]
  \centering
   \resizebox{1.0\textwidth}{!}{%
  \begin{tabular}{lllllllllllllllll}
  \bf Model & \bf En & \bf Fr & \bf Es & \bf De & \bf El & \bf Bg & \bf Ru & \bf Tr & \bf Ar & \bf Vi & \bf Th&  \bf Zh & \bf Hi & \bf Sw & \bf Ur & \bf Ave.  \\
   \hline
  \hline
  mBERT & 81.3 & 73.4 & 74.3 & 70.5 & 66.9 & 68.2 & 68.5 & 59.5 & 64.3 & \bf 70.6 & 50.7 & 68.8 & 59.3 & 49.4 & 57.5 & 65.5 \\
  Ours & \bf 81.5 & \bf 74.1* & \bf 74.9* & \bf 71.2* & \bf 67.1 & \bf 68.7* & \bf 68.6 & \bf 61.0* & \bf 66.2* & 70.5 & \bf 53.8* & \bf 69.1 & \bf 59.8* & \bf 50.6* & \bf 58.6* & \bf 66.4* \\
    \end{tabular}
    }
    \caption{\label{tab:cross} Results of mBERT and our fine-tuned model on XNLI~\citep{conneau2018xnli}. Our objectives can improve the model cross-lingual transfer ability. ``*'' denotes significant differences using paired bootstrapping (p$<$0.05) .
    }
  \end{table*}

\begin{figure*}[t]
\centering
\includegraphics[width=1.0\textwidth]{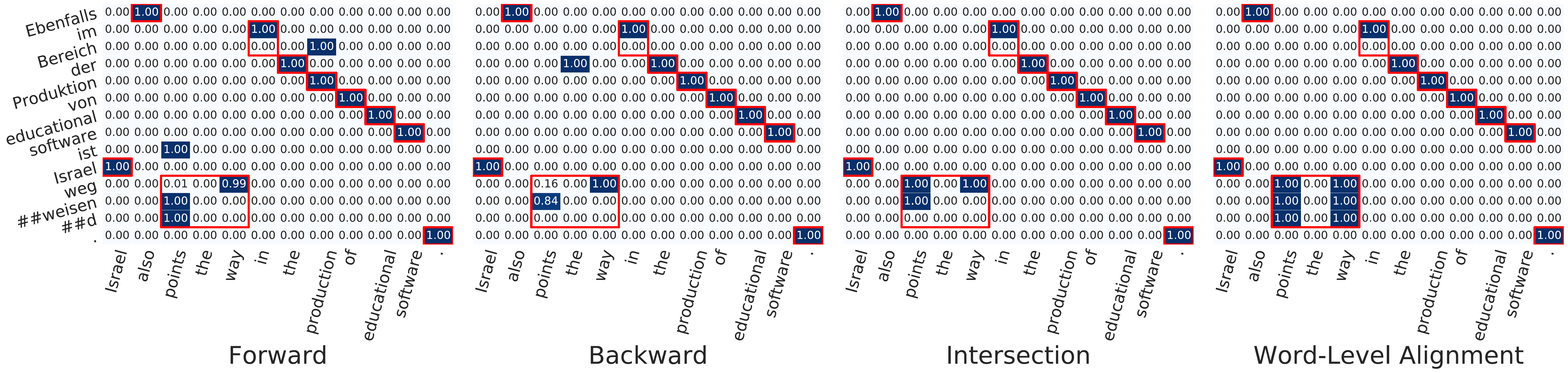}
\caption{ An example of extracting alignments from our fine-tuned model using \textit{softmax}. Red boxes indicate the gold alignments. The fine-tuned model can generate more accurate alignments then vanilla mBERT (Figure~\ref{fig:example1}).}
\label{fig:example}
\end{figure*}  

\paragraph{Annotation Projection.} Word alignment has been a useful tool in cross-lingual annotation projection~\citep{yarowsky2001inducing,nicolai-yarowsky-2019-learning}. Therefore, it would be interesting to see if our model can be beneficial in these settings. To this end, we evaluate our model and baselines on cross-lingual named entity recognition (NER). We train a BERT-based NER model on the CoNLL 2003 English data~\citep{conll2003} and test it on the CoNLL 2002 Spanish data~\citep{conll2002}. We use Google Translate to translate Spanish test set into English, predict the labels using the NER model, then project the labels from English to Spanish using word aligners. From Table~\ref{tab:ner}, we can see that our model is also better than baselines in this setting, demonstrating its usefulness in cross-lingual annotation projection.

\paragraph{Sentence-Level Representation Transfer.} We also test if the aligned representations are beneficial for sentence-level cross-lingual transfer. In doing so, we perform experiments on XNLI~\citep{conneau2018xnli}, which evaluates cross-lingual sentence representations in 15 languages on the task of natural language inference (NLI). We train our models with the provided 10k parallel data on the 15 languages, fine-tune our model on the English NLI data, then test its performance on other languages. As shown in Table~\ref{tab:cross}, our model can outperform the baseline, indicating the aligned word representations can also be helpful for sentence-level cross-lingual transfer.

\paragraph{Alignment Examples.} We also conduct qualitative analyses as shown in Figure~\ref{fig:1},~\ref{fig:example1} and~\ref{fig:example}. After fine-tuning, the learned contextualized representations are more aligned, as the cosine distances between semantically similar words become closer, and the extracted alignments are more accurate. More examples are shown in Appendix~\ref{app:more}.

\section{Related Work}

Based on the IBM translation models~\citep{brown1993mathematics}, many statistical word aligners have been proposed~\citep{vogel1996hmm,ostling2016efficient}, including the current most popular tools GIZA++~\citep{och2000improved,och2003systematic,gao2008parallel} and fast\_align~\citep{dyer2013simple}.

Recently, there is a resurgence of interest in neural word alignment~\citep{tamura2014recurrent,alkhouli2018alignment}. Based on NMT models trained on parallel corpora, researchers have proposed several methods to extract alignments from them~\citep{luong2015effective,zenkel2019adding,garg2019jointly,li2019word} and successfully build an end-to-end neural model that can outperform statistical tools~\citep{zenkel2020end}. However, there is an inherent discrepancy between translation and word alignment: translation models are directional and the source and target side are treated differently, while word alignment is a non-directional task. Therefore, certain adaptations are required for translation models to perform word alignment.

Another disadvantage of MT-based word aligners is that they cannot easily utilize contextualized embeddings. Using learned representations to improve word alignment have been investigated~\citep{sabet2016improving,pourdamghani2018using}. Recently, pre-trained LMs~\citep{peters2018deep,Devlin2019BERTPO,Brown2020LanguageMA} have proven to be useful in cross-lingual transfer~\citep{libovicky2019language,hu20icml}. In word alignment,~\citet{sabet2020simalign} propose effective methods to extract alignments from multilingual LMs without explicit training on parallel data.
In this work, we propose better alignment extraction methods and combine the best of the two worlds by fine-tuning contextualized embeddings on parallel data.

There are also work on supervised neural word alignment~\citep{stengel2019discriminative,nagata2020supervised}. However, supervised data are not always accessible, making their methods inapplicable in many scenarios. In this paper, we demonstrate that our model can incorporate supervised signals if available and perform semi-supervised learning, which is a more realistic and general setting.

Some work on bilingual lexicon induction also share similar general ideas with ours. For example,~\citet{zhang2017earth} minimize the earth mover’s distance to match the embedding distributions from different languages. Similarly,~\citet{grave2019unsupervised} present an algorithm to align point clouds with Procrustes~\citep{schonemann1966generalized} in Wasserstein distance for unsupervised embedding alignment.
\section{Discussion and Conclusion}
We present a neural word aligner that achieves state-of-the-art performance on five diverse language pairs and obtains robust performance in zero-shot settings. We propose to fine-tune multilingual embeddings with objectives suitable for word alignment and develop two alignment extraction methods. We also demonstrate its applications in semi-supervised settings. We hope our word aligner can be a tool that can be used out-of-the-box with good performance over various language pairs. Future directions include designing better training objectives and experimenting on more language pairs.

Also, note that we mainly evaluate our word aligners using AER following previous work, which has certain limitations. For example, it may not be well-correlated with statistical machine translation performance~\citet{fraser2007measuring} and different types of alignments can be suitable for different tasks or conditions~\citep{lambert2012types,stymne2014estimating}. Although we have evaluated models in annotation projection and cross-lingual transfer settings, alternative metrics~\citep{tiedemann2005optimization,sogaard2009empirical,ahrenberg2010alignment} are also worth considering in the future.

\section*{Acknowledgement}
We thank our reviewers for helpful suggestions.
\bibliography{eacl2021}
\bibliographystyle{acl_natbib}

\clearpage
\appendix

\section{Implementation Details}

 We use the AdamW optimizer~\citep{Loshchilov2019DecoupledWD} with a learning rate of 2e-5 and the batch size is set to 8. Following~\citet{peters2019sparse}, we set $\alpha$ to 1.5 for $\alpha$-entmax. The threshold $c$ is set to 0 for $\alpha$-entmax and 0.001 for \textit{softmax} and optimal transport. 
 We train our models on one 2080 Ti for one epoch and it takes 3 to 24 hours for the model to converge depending on the size of the dataset.
 We evaluate the model performance using Alignment Error Rate (AER).

\section{Analysis}
In this section, we conduct more analyses of our models.

\paragraph{Monolingual Alignment.}  We also investigate how our models perform in monolingual alignment settings. Previous methods~\cite{maccartney2008phrase,yao2013lightweight,yao2013semi,sultan2014back} typically exploit external resources such as WordNet to tackle the problem. As shown in Table~\ref{tab:mono}, mBERT can outperform previous methods in terms of recall and F$_1$ without any fine-tuning. Our multilingually fine-tuned model can achieve better recall and slightly better F$_1$ score than the vanilla mBERT model, and fine-tuning our model with supervised signals can achieve further improvements. 

\begin{table}[t]
  \centering
   \resizebox{0.4\textwidth}{!}{%
  \begin{tabular}{llll}
  \bf Model & \bf Prec. \% &  \bf Rec.\% & \bf F$_1$ \%   \\
   \hline
  \hline
  \multicolumn{4}{l}{ \it Baseline} \\
  \hline
 ~\citet{yao2013lightweight}  &91.3& 82.0 & 86.4 \\
 ~\citet{yao2013semi}& 90.4 & 81.9 & 85.9 \\
 \citet{sultan2014back} & \bf 93.5 & 82.6 & 87.6 \\
   \hline
  \hline
  \multicolumn{4}{l}{ \it Ours} \\
  \hline
  \multirow{1}{*}{mBERT} & 87.0 & 89.0 & 88.0  \\
  \multirow{1}{*}{Ours-Multilingual} & 87.0 & 89.3 & 88.1  \\
  Ours-Supervised & 87.2 &  \bf 89.8 &  \bf 88.5 \\
    \end{tabular}
    }
    \caption{\label{tab:mono} Our model is also effective in monolingual alignment settings.}
  \end{table}

\paragraph{Sensitivity Analysis.} We also conduct a sensitivity analysis on the threshold $c$ for our \textit{softmax} alignment extraction method. As shown in Table~\ref{tab:sensitivity}, our method is relatively robust to this threshold. In particular, after fine-tuning, the AERs change within 0.5\% when varying the threshold. 

\begin{table}[t]
  \centering
   \resizebox{0.5\textwidth}{!}{%
  \begin{tabular}{llllllll}
  \bf Model &  $\textbf{c}$. &\multicolumn{1}{c}{ \bf De-En} & \bf Fr-En & \bf Ro-En & \bf Ja-En & \bf Zh-En  \\
   \hline
  \hline
  \multirow{7}{*}{mBERT} 
  & 1e-6 & 17.3 & 6.0 & 27.2 & 45.2 & 18.9\\
  & 1e-5 & 17.3 & 5.9 & 27.4 & 45.1 & 18.6 \\
  & 1e-4 & 17.3 & 5.7 & 27.6 & 45.3 & 18.3 \\
  & 1e-3 & 17.4 & 5.6 & 27.9 & 45.6 & 18.1 \\
  & 1e-2 & 17.7 & 5.6 & 28.4 & 45.8 & 18.2 \\
  & 1e-1 & 18.1 & 5.6 & 28.9 & 46.3 & 18.3 \\
  & 5e-1 & 18.4 & 5.6 & 29.5 & 47.0 & 18.7 \\
  \hline 
  \hline 
  \multirow{7}{*}{Ours-Multilingual} 
  & 1e-6 & 15.4 & 4.6 & 22.7 & 38.2 & 14.1 \\
  & 1e-5 & 15.4 & 4.5 & 22.7 & 38.1 & 14.0 \\
  & 1e-4 & 15.3 & 4.5 & 22.6 & 37.9 & 13.9 \\
  & 1e-3 & 15.3 & 4.4 & 22.6 & 37.9 & 13.8 \\
  & 1e-2 & 15.3 & 4.3 & 22.7 & 37.9 & 13.8 \\
  & 1e-1 & 15.4 & 4.3 & 22.8 & 38.0 & 13.8 \\
  & 5e-1 & 15.4 & 4.2 & 23.0 & 38.2 & 13.9 \\
    \end{tabular}
    }
    \caption{\label{tab:sensitivity} Our \textit{softmax} alignment extraction method is relatively robust to the threshold $c$. }
  \end{table}

\begin{figure*}[t]
\centering
\subfloat[mBERT Itermax]{
\includegraphics[width=1.0\textwidth]{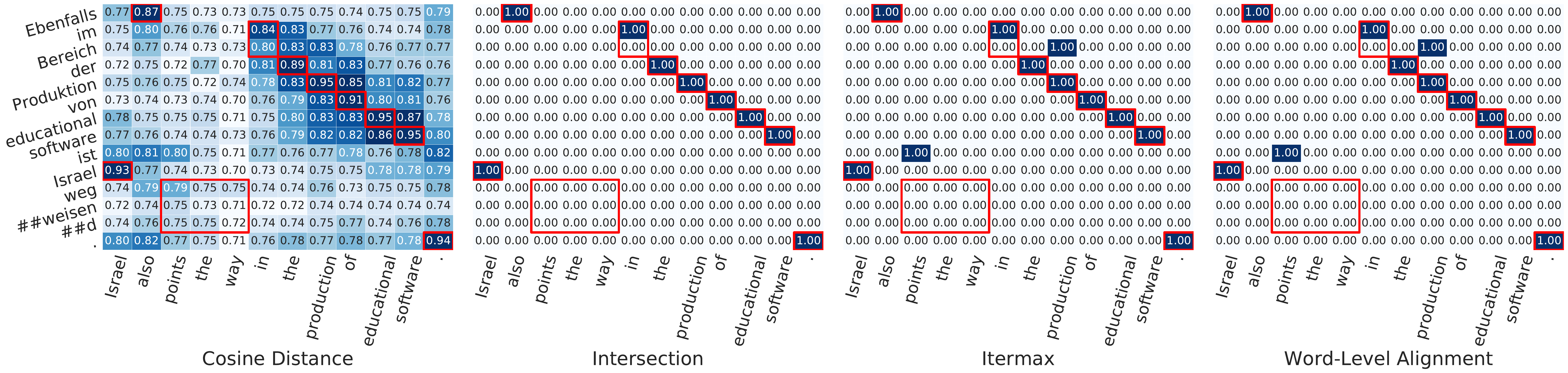}
\label{fig:sub1}
} \hfill
\subfloat[mBERT \textit{softmax}]{
\includegraphics[width=1.0\textwidth]{base_softmax2.pdf}
\label{fig:sub2}
} \hfill
\subfloat[Fine-tuned IterMax]{
\includegraphics[width=1.0\textwidth]{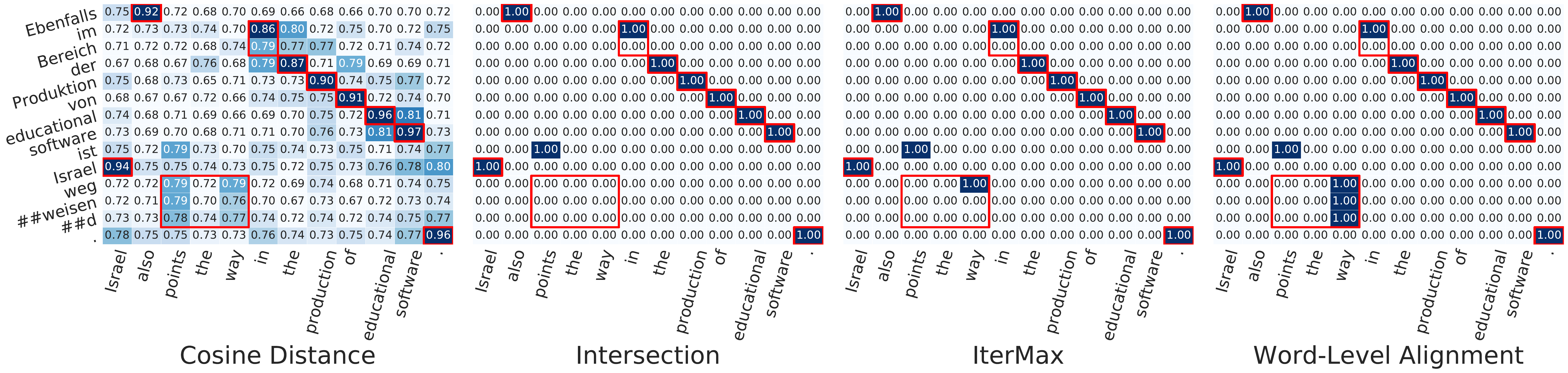}
\label{fig:sub3}
} \hfill
\subfloat[Fine-tuned \textit{softmax}]{
\includegraphics[width=1.0\textwidth]{ours_softmax2.pdf}
\label{fig:sub4}
}
\caption{ Extracting alignments from our model using IterMax\cite{sabet2020simalign} and our \textit{softmax} method from the vanilla and fine-tuned mBERT models.}
\label{fig:example4}
\end{figure*}  

\paragraph{Comparisons with IterMax.} IterMax is the best alignment extraction method in SimAlign~\cite{sabet2020simalign}. The results in the main paper have demonstrated that our alignment extraction methods are able to outperform IterMax. In Figure~\ref{fig:example4}, we can see that the IterMax algorithm tends to sacrifice precision for a small improvements in recall, while our model can generate more accurate alignments.

\begin{table*}[ht]
  \centering
   \resizebox{0.8\textwidth}{!}{%
  \begin{tabular}{lllllll}
 \multicolumn{1}{l}{ \bf Model }& \bf Objective &\multicolumn{1}{c}{ \bf De-En} & \bf Fr-En & \bf Ro-En & \bf Ja-En & \bf Zh-En   \\
   \hline
  \hline
  \multicolumn{7}{l}{ \it Ours-Bilingual} \\
  \hline
  \multirow{6}{*}{$\alpha$-entmax } 
   & \it All  & 16.1 &    4.1 & 23.4 & 38.6 & 15.4\\
   \cdashline{2-7}
     & \it All w/o MLM & 15.6 & 4.2 & 23.3 & 38.8 & 15.1\\
   & \it All w/o TLM  & 16.4 & 4.3 & 23.7 & 40.1 & 15.3\\
   & \it All w/o SO  & 17.8 &  4.7 & 23.9 & 39.4 & 16.3\\
 & \it All w/o PSI & 16.5 & 4.2 & 23.1 & 38.5 & 15.4 \\
  \hline
   \multirow{6}{*}{\it softmax } 
      & \it All  &  15.6 & 4.4 &  23.0 &   38.4 &   15.3\\
       \cdashline{2-7}
   & \it All w/o MLM  &  15.5 & 4.2 & 23.2 & 38.9 & 14.9\\
   & \it All w/o TLM  & 15.9 & 4.5 & 23.7 & 40.1 & 15.1 \\
   & \it All w/o SO  & 17.4  & 4.7 & 23.2 & 38.6 & 16.3\\
  & \it All w/o PSI  &  15.6 &4.3  & 23.1 &38.8 & 15.4\\
   \hline
  \hline
  \multicolumn{7}{l}{ \it Ours-Multilingual} \\
  \hline
  \multirow{5}{*}{$\alpha$-entmax } 
   & \it All & 15.4 &  4.1 & 22.9 &  37.4 & 13.9 \\
   \cdashline{2-7}
  & \it  All w/o MLM & 15.1 & 4.2 & 22.8 & 37.8 & 13.7  \\
  & \it  All w/o TLM & 16.4 & 4.4 & 23.3 & 39.7 & 14.4\\
  & \it  All w/o SO  & 17.5 & 4.6 & 23.6 & 40.0 & 15.6\\
  & \it  All w/o PSI & 15.5 & 3.9 & 23.0 & 38.2 & 14.1 \\
  \hline
  \multirow{5}{*}{\it softmax} 
  & \it All &   15.3 & 4.4 & 22.6 & 37.9 & 13.6 \\
   \cdashline{2-7}
    &  \it All w/o MLM & 15.3 & 4.4 & 22.8 & 38.6 & 13.7\\
   &  \it All w/o TLM & 15.5 & 4.7 & 22.9 & 39.7 & 14.0 \\
   & \it All w/o SO & 16.9 & 4.8 & 23.0 & 39.1 & 15.4 \\
  & \it  All w/o PSI &  15.4 & 4.4 & 22.7 & 37.9 & 13.8 \\
 
  \hline
    \end{tabular}
    }
    \caption{\label{tab:trash} Ablation studies on training objectives.}
  \end{table*}

\paragraph{Ablation Studies on Training Objectives.} Table~\ref{tab:trash} presents more ablation studies on our training objectives. We can see that the self training objective is the most effective one, with the translation language modeling objective being the second and the parallel sentence identification objective being the third. The masked language modeling objective can sometimes hurt the model performance, possibly because of the translation language modeling objective.
\paragraph{Experiments on More Language Pairs.} We also test our alignment extraction methods on other language pairs following the setting of~\citet{sabet2020simalign} without fine-tuning as shown in Table~\ref{tab:lang}.\footnote{Their English-Persian dataset is unavailable at the time of writing the paper.}

\begin{table}[t]
  \centering
   \resizebox{0.4\textwidth}{!}{%
  \begin{tabular}{llll}
  \bf Model & \bf En-Cs  &  \bf En-Hi    \\
   \hline
  \hline
 GIZA++ & 18.2 & 51.8 \\
 SimAlign & 13.4 & 40.2 \\
  Ours ({\it softmax}, $c$=1e-3) &  \bf 12.3 & 41.2 \\
   Ours ({\it softmax}, $c$=1e-5) &   12.7 &   39.5 \\
    Ours ({\it softmax}, $c$=1e-7) &  13.3 & \bf  39.2 \\
    \end{tabular}
    }
    \caption{\label{tab:lang} Performance on more language pairs.}
  \end{table}
  
\begin{figure*}[t]
\centering
\includegraphics[width=0.9\textwidth]{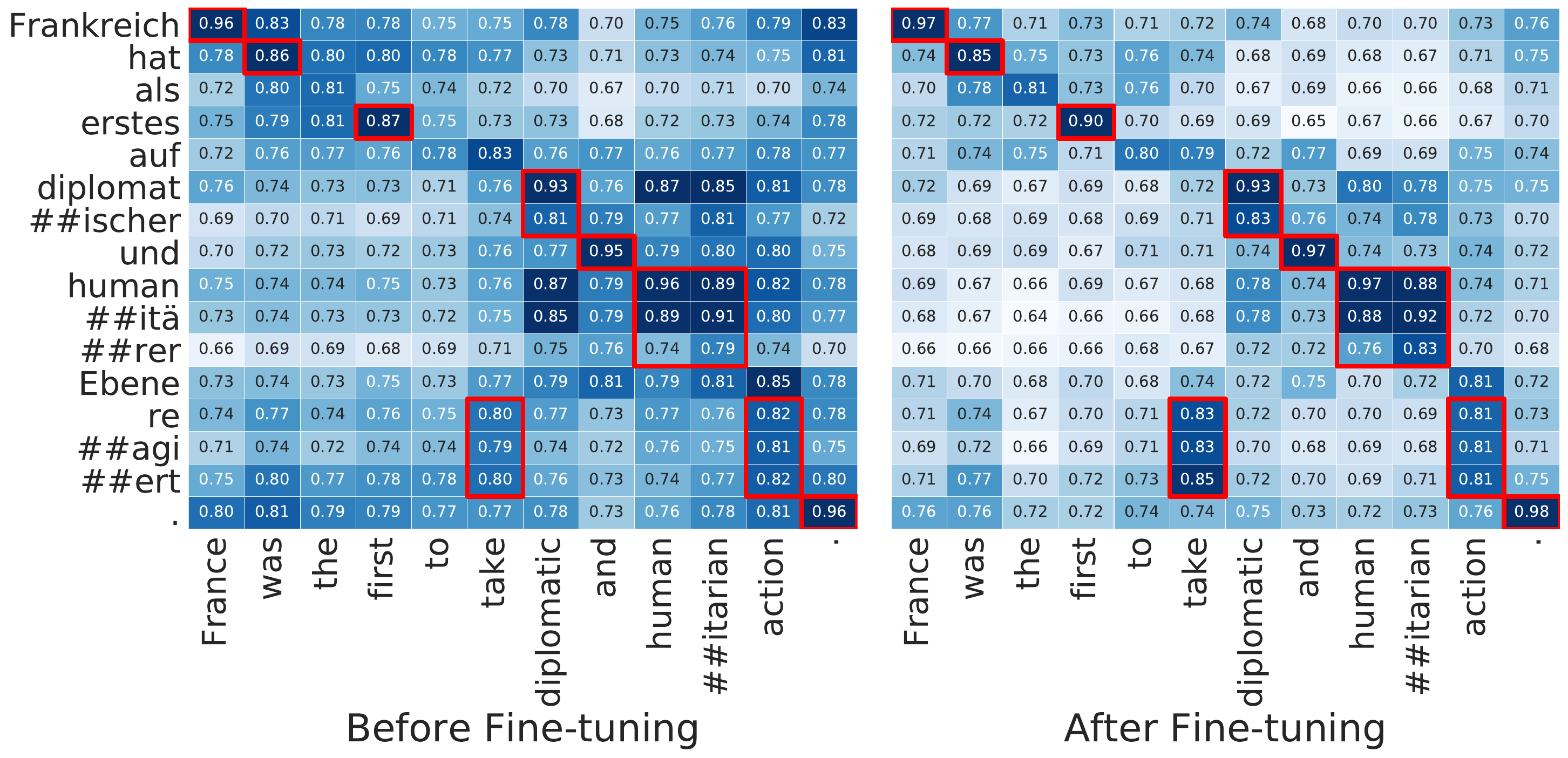} \\
\includegraphics[width=0.9\textwidth]{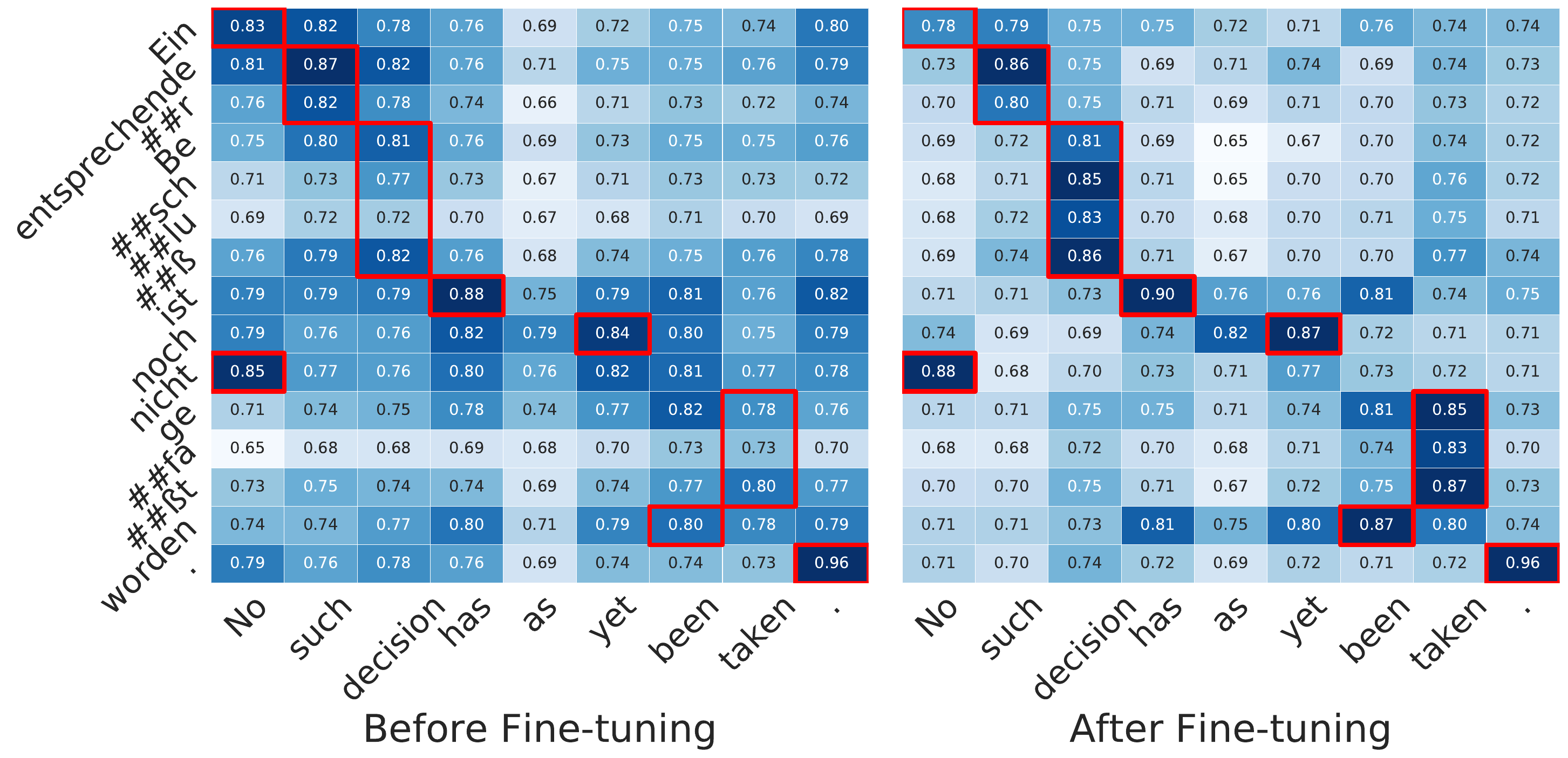} \\
\includegraphics[width=0.9\textwidth]{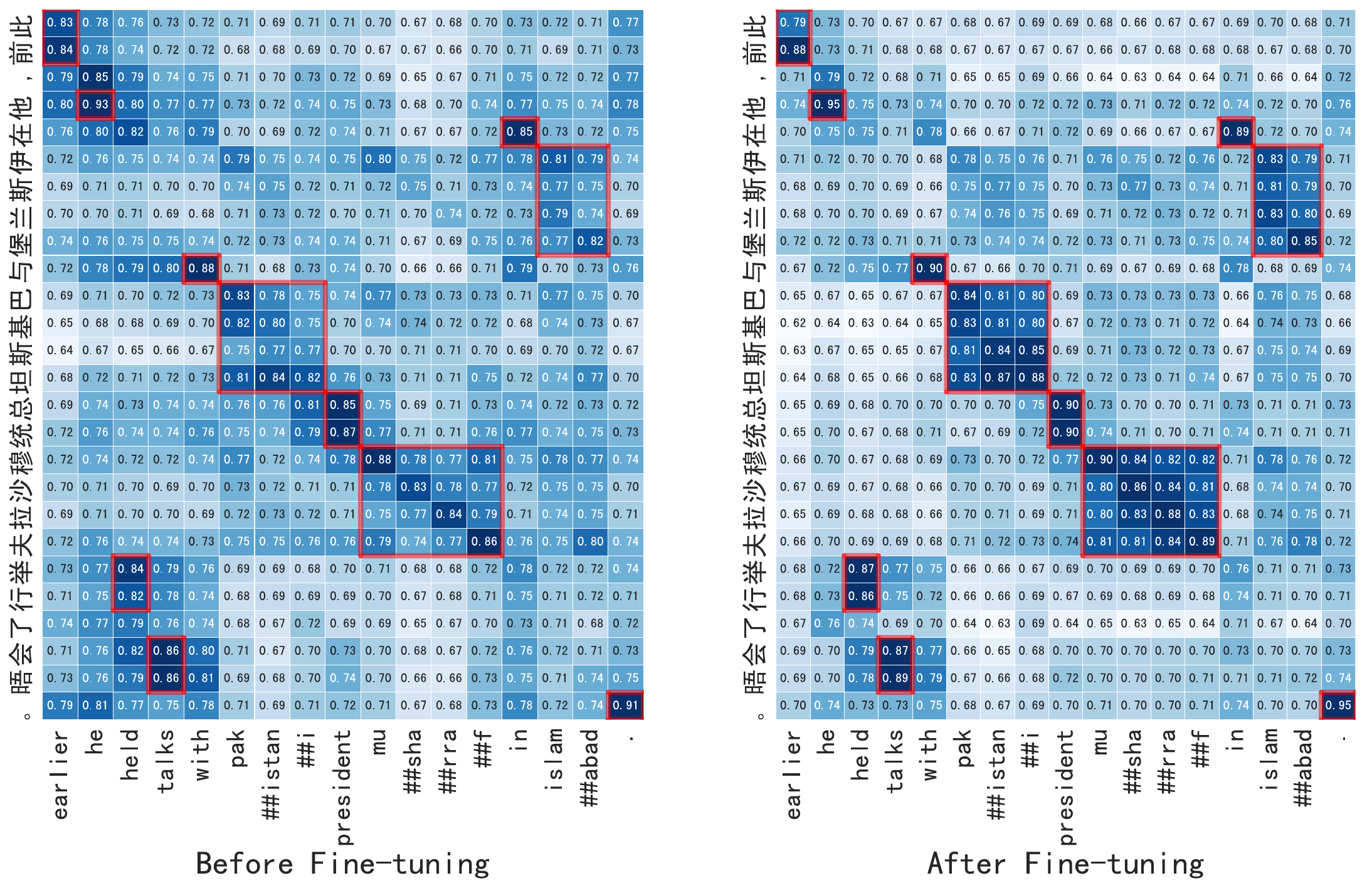}
\caption{Cosine similarities between subword representations in a parallel sentence pair before and after fine-tuning. Red boxes indicate the gold alignments.} 
\label{fig:more}
\end{figure*}

\label{app:more}
\paragraph{More Qualitative Examples.} In addition to the examples provided in the main text, we also present some randomly sampled samples in Figure~\ref{fig:more}. We can clearly see that our model learns more aligned representations than the baseline model.


\end{document}